\documentclass[journal,article,accept,pdftex,moreauthors]{Definitions/mdpi}

\firstpage{1} 
\pubvolume{1}
\issuenum{1}
\articlenumber{0}
\pubyear{2024}
\copyrightyear{2024}
\datereceived{ } 
\daterevised{ } 
\dateaccepted{ } 
\datepublished{ } 
\hreflink{https://doi.org/} 

\usepackage{graphics} 
\usepackage{epsfig} 
\usepackage{mathptmx} 
\usepackage{times} 
\usepackage{amsmath} 
\usepackage{amssymb}  

\newtheorem{Proof}{Proof}

\Title{Discrete-Time Stress Matrix-Based Formation Control of General Linear Multi-Agent Systems}

\TitleCitation{Discrete-Time Stress Matrix-Based Formation Control of General Linear Multi-Agent Systems}

\Author{Okechi Onuoha $^{1}$, Suleiman Kurawa $^{2}$, Zezhi Tang $^{3}$ and Yi Dong $^{4}$}

\AuthorNames{Okechi Onuoha, Suleiman Kurawa, Zezhi Tang and Yi Dong}

\AuthorCitation{Onuoha, O.; Kurawa, S.; Tang, Z.; Dong, Y.}

\address{%
$^{1}$ \quad School of Computing, Engineering, and Intelligent Systems,
		Ulster University; o.onuoha@ulster.ac.uk\\
$^{2}$ \quad School of Electrical and Electronic Engineering, 
		The University of Manchester; suleiman.kurawa@postgrad.manchester.ac.uk\\
$^{3}$ \quad Department of Automatic Control and Systems Engineering, University of Sheffield; zezhi.tang@sheffield.ac.uk\\
$^{4}$ \quad School of Electronics \& Computer Science, University of Southampton; yi.dong@soton.ac.uk}

\abstract{
This paper considers the distributed leader-follower stress-matrix-based affine formation control problem of discrete-time linear multi-agent systems with static and dynamic leaders. In leader-follower multi-agent formation control, the aim is to drive a set of agents comprising leaders and followers to form any desired geometric pattern and simultaneously execute any required manoeuvre by controlling only a few agents denoted as leaders. Existing works in literature are mostly limited to the cases where the agents' inter-agent communications are either in the continuous-time settings or the sampled-data cases where the leaders are constrained to constant (or zero) velocities or accelerations. Here, we relax these constraints and study the discrete-time cases where the leaders can have stationary or time-varying velocities. We propose control laws in the study of different situations and provide some sufficient conditions to guarantee the overall system stability. Simulation study is used to demonstrate the efficacy of our proposed control laws.}


\begin{document}




\section{INTRODUCTION}	

Formation control of multi-agent systems (MASs) has attracted several research interests in recent times because of its wide applications in both civil and military operations. This includes areas such as environment and facilities monitoring using e.g., unmanned aerial vehicles (UAVs); military attack and defence systems using multiple unmanned aircraft, boats, and underwater vehicles; and agricultural applications such pest control using UAVs. Some other application areas includes mobile robots control and sensor networks. The primary goal of formation control is to steer a team of agents to form any required geometric pattern and also to simultaneously accomplish any required manoeuvre control of the team such as formation translation, scaling, shearing, rotation, etc.

Researcher have used different strategies to accomplish formation in the last two decades. For example, in the last decade, the consensus-based formation control strategy has been used to study formation control problems in \cite{ding}, \cite{syr5} and \cite{syr6}. The consensus-based strategy primarily consists of three methods namely the: displacement-based \cite{syr8}, distance-based \cite{syr9}, and bearing-based \cite{syr10} formation control methods. They accomplish formation control by imposing pre-defined constant off-sets on a chosen reference. The reference can be for example, the leader position or consensus value of the agents. The imposed constant off-set can be on the agents bearing (for bearing-based formation control), distance (for distance-based formation control), or displacement (for displacement-based formation control). A major problem with these methods of formation control is that the imposed off-sets are defined at the design stage and normally cannot be changed during operation without a redesign of the system. This therefore limits the formation manoeuvre the agents can accomplish. Note that agents of these methods cannot simultaneously accomplish the commonly required formation manoeuvres such as scaling, rotation, translation, shearing, etc \cite{me_jfi}. Recently, the complex Laplacian \cite{syr20}, \cite{syr21} formation control strategy has been used to overcome the limitation of the consensus-based strategy. The strategy is able to accomplish most of the commonly required formation manoeuvre control. Unfortunately, the strategy is limited to 2-dimensional systems, e.g. formation control on a 2-dimensional plane surface. This is considered very restrictive. This shortfall inspired the stress-matrix-based formation control strategy.

Very recently, the stress-matrix-based formation control strategy has been identified and used for formation control because of its promising prospects. This is primarily because it is able to accomplish most of the commonly required manoeuvres in any dimension. This is the main advantage over the previous strategies. Previous studies using the stress-matrix-based strategy have been mostly limited to the continuous-time settings, e.g. \cite{wang_us}, \cite{f1}, and \cite{f2}. Exceptions to this are the works of \cite{me_us}, \cite{me_acc} which considered the sampled-data settings. They study the sampled-data cases where the agents are modelled using triple-integrator agent dynamics with the assumption that the agents have constant/static jerks. This can be very restrictive.

In this paper, we relax this restrictive requirement by studying the sampled-data case where the velocities of the agents are allowed to dynamically change. The results can be extended to consider double- and triple-integrators agent dynamics where the agents respective accelerations and jerks are allowed to dynamically change. Here, we first study the case where the agents velocities are constant (or slowly changing or zero). We then consider the case with dynamic velocities. Two control laws are used to study the different cases. We then proceed to the general linear discrete-time case. Sufficient conditions are provided to guarantee the overall stability of the system. With the proposed control laws, sampled-data affine formation control of multi-agent systems with stationary and dynamic targets can be accomplished. A simulation study is used to verify the efficacy of the proposed control law.

The rest of this paper is organised as follows. 

\section{PRELIMINARIES}	

This section introduces some preliminary notations and results used in the rest of this paper.
\subsection{Basic Graph Notations}

Consider a collection of $n$ agents that are required to cooperate to execute some tasks. The inter-agent communications are modelled using a communication a graph  $\mathcal{G} (\mathcal{V}, \mathcal{E})$, which primarily comprise of vertices $\mathcal{V}$ (also called nodes) and edges $\mathcal{E}$. A direct communication path between two neighbouring vertices are denoted using an edge. For example, assuming that node $i$ is capable of sending information to a neighbouring node $j$ (or node $j$ is allowed to obtain information from node $i$), then the edge ($i,j$) is considered to exist, i.e., $(i,j)\in\mathcal{E}$. The weight of the edge $(i,j)$ is denoted using $w_{ji}$ if it exists and $0$ if does not. If for every given edge $(i,j) \in \mathcal{E}$; there also exist the edge $(j,i) \in \mathcal{E}$, the communication graph is said to be undirected, but if otherwise (even for a single edge in the entire graph), the graph is said to be directed \cite{z8}. The set of nodes a given node can receive information from are considered to be its neighbours. The notation $\mathcal{N}_i$ is used to denote the set of neighbours of node $i$. In this paper, it is assumed that each agent (or node) knows its own state and also those of its neighbours. The notation $\otimes$ is used to denote the Kronecker product.

The term \textit{configuration} is used to denote a collection of nodes defined by their positions in Euclidean coordinate space, $\mathbb{R} ^d$. This is given by $p=[p_1^T,...,p^T_n]^T$ where  $p_i \in \mathbb{R} ^d $. The term \textit{framework} in $\mathbb{R}^d$, defined by  $\mathcal{F}=(\mathcal{G},p)$, is a communication graph defined with its configuration. Two frameworks, $(\mathcal{G},p)$ and $(\mathcal{G},q)$ are considered to be equivalent, written as $(\mathcal{G},p) \equiv (\mathcal{G},q)$, if

\begin{equation*}
	\parallel p_i - p_j \parallel = \parallel q_i - q_j \parallel, \quad \forall (i,j) \in \mathcal{E}.
\end{equation*}
The two frameworks, $(\mathcal{G},p)$ and $\mathcal{G}(q)$, are considered to be congruent, written as $(\mathcal{G},p) \cong (\mathcal{G},q)$, if 
\begin{equation*}
	\parallel p_i - p_j \parallel = \parallel q_i - q_j \parallel, \quad \forall {i,j} \in \mathcal{V}.
\end{equation*}
A given framework in $\mathbb{R} ^d$ is considered to be globally rigid if $(\mathcal{G},p) \equiv (\mathcal{G},q)$ infers that $(\mathcal{G},p) \cong (\mathcal{G},q)$. This implies that any framework in $\mathbb{R} ^d$ that is considered to be equivalent to $(\mathcal{G},p)$ is congruent to it also. A configuration, $p$  is considered to be universally rigid if for  any $\mathbb{R}^{d_1}$, where $d_1$ denotes any defined positive integer, $(\mathcal{G},p) \equiv (\mathcal{G},q)$ infers $(\mathcal{G},p) \cong (\mathcal{G},q)$. Therefore, universal rigidity naturally implies global rigidity but global rigidity does not confer a universal rigidity status \cite{a3}, \cite{R_Connelly} and \cite{syr22}.

\subsection{Affine Span}

The affine span, $\mathcal{S}$, of a given set of points, $\{p_i\}^n_{i=1} \in \mathbb{R} ^d$ is defined by

\begin{equation*}
	\mathcal{S}=  \left\{\sum\limits ^n_{i=1} a_ip_i ~ : ~ a_i \in \mathbb{R} ~ \forall i ~~ \mbox{and} ~ \sum\limits ^n_{i=1}a_i = 1\right\}.
\end{equation*}

\noindent
To span a $d$-dimensional space affinely, $d+1$ affinely independent points are required. The affine span of any two unique points is a line connecting both points. Similarly, the affine span of three distinct points that are not collinear is a $2$-dimensional plane passing through the three points. Higher dimensions follow the analogy.

A defined affine span can be translated so that it contains the origin, and therefore a linear space with a dimension that is the same as the affine space. Thus, given any $d$-dimensional affine span, one can say that the points affinely span $\mathbb{R}^d$. In this paper, we consider the case where $d+1$ leaders whose span is $\mathbb{R}^d$ is chosen for the affine formation control in defined $d$-dimensional space.

\subsection{Stress Matrix}

Consider an undirected graph having a framework, $\mathcal{F} =(\mathcal{G},p)$, the associated stress is defined as a collection of weights, $w_{ij}$, assigned to the respective edges. The stress satisfying

\begin{equation}
	-\sum w_{ij} (p_i - p_j)= ~ 0, \quad i \in \mathcal{V},
	\label{basicEdn1}
\end{equation}
\noindent
is considered to be an equilibrium stress \cite{a3}, \cite{R_Connelly}. Using matrix notation, Equation (\ref{basicEdn1}) can be rewritten as 

\begin{equation*}
	-( \Omega \otimes I_d)p= ~ 0,
\end{equation*}
where $\Omega \in \mathbb{R}^{n \times n}$ denotes the stress matrix, and its entries are given by

\begin{align}
	\Omega_{ij} = 
	\begin{cases}
		\sum\limits_{j \in \mathcal{N}_i} w_{ij(k)} ,  & \quad \text{for } i=j,  \\
		-w_{ij},      & \quad \text{for } i \ne j, ~ ~ (j,i) \in \mathcal{E},   \\
		0,  & \quad \text{for } i \ne j, ~ ~ (j,i) \notin \mathcal{E}. \\
	\end{cases}
	\label{stress}
\end{align}

\noindent
It may be worth noting that the traditional Laplacian matrices and the stress matrices of graphs have some similar properties, with a major difference being that in the stress matrix, the off-diagonal entries can be positive or negative values (including zero), unlike in the Laplacian matrices whose off-diagonal entries are only allowed to be negative values (or zero). For convenience, it is common to partition $\Omega$ as 
$$
\Omega =  \begin{bmatrix}
	\Omega_{ll} & \Omega_{lf}\\
	\Omega_{fl} & \Omega_{ff}
\end{bmatrix}.
$$
where the sub-matrices $\Omega_{ll}$ and $\Omega_{ff}$ are respectively $n_l \times n_l$ and $n_f \times n_f$.
Note that the values of all $w_{ij}$ are to be computed.

\subsection{Stress Matrix Design}  \label{subsection stress matrix design}

In affine formation control, stress-matrix plays a key role . For a defined affine formation to be stabilizable, the underlining framework $(\mathcal{G},p)$ is required to be universally rigid. It is the universal rigidity of a framework that guarantees its uniqueness in every dimensions. Therefore, to ensure the accomplishment of a required affine formation control, it is important that the underlining framework is unique. This implies that the framework needs to be rigid (universally) to ensure proper stabilization. The required universal rigidity is normally accomplished via suitable stress matrix design. Note that, a framework, $\mathcal{F} =(\mathcal{G},p)$, is used to denote a communication graph defined together with the positions of its nodes. Universal rigidity is a more demanding rigidity than global rigidity. Global rigidity only requires the uniqueness of a given framework in the entire space of the given dimension. Unfortunately, necessary and sufficient conditions to guaranteeing universal (or global) rigidity is currently still lacking in existing literature to the best of our knowledge.

Analysis and designs primarily concerned with multi-agent systems coordination commonly focus on the special universal (and global rigidity) cases where the underlining framework has a generic configuration. A configuration is said to be generic if its coordinates are algebraically independent over the integer \cite{a2,a4}. Sufficient conditions to guarantee universal (or global) rigidity exist in existing literature. The design of stress-matrix is outside the scope of this readers interested in further details can see e.g., \cite{a3,a2,a4}. Also, see \cite{me_acc, me_jfi} for further details. Some key requirements/highlights are presented as follows.

\begin{Lemma} \label{Lemma_Graphical_Rigidity_Requirement} \cite{a2,a3,a4}:
	Given a framework ($ \mathcal{G},p$) with an undirected communication graph, a generic configuration in $\mathbb{R} ^d$ and number of nodes $n \geq d+2 $. The framework is universally rigid only if its communication graph is $(d+1)$-connected and has a stress matrix, $\Omega$ that has a rank of $n-d-1$ and is positive semi-definite. 
\end{Lemma}

\begin{Assumption} \label{Assumption stress matrix}
	The considered framework $(\mathcal{G},p)$ is \textit{generically universally rigid}. 
\end{Assumption}

\begin{Remark}
	Assumption \ref{Assumption stress matrix} ensures that the rank$(\Omega)$ of the stress matrix is  $n-d-1$.
\end{Remark}

\begin{Assumption} \label{Assumption positive omega ff}\cite{PosEig}  \label{lemma positive eigenvalues}
	Assume Assumption \ref{Assumption stress matrix} hold, then the eigenvalues of $\Omega_{ff}$ all have positive real parts. 
\end{Assumption}
See \cite{syr_shiyu,constructing_rigid_formation,me_us,me_acc} for more details.


\subsection{Leader Selection and Affine Localizability}

To effectively use the leaders to manipulate the entire agent formation, it is important to carefully select the leaders. This subsection presents a guide on how to select the leaders for proper affine formation control.

\begin{Lemma} \label{Affine_Localability} \cite{syr_shiyu}
	Assume that the nominal formation for the framework $(\mathcal{G},p)$ consist of $n_l$ leaders and $n_f$ followers. The target positions of all the followers $p_f^*$ can be uniquely derived from  
	\begin{equation*}
		p_f^* = -\Omega_{ff}^{-1}\Omega_{fl} p_l^*
	\end{equation*}
	for all $p=[p_l^T, p_f^T]^T$ belonging to the set of affine transformations of the reference (nominal) formation, if $\Omega_{ff}$ is nonsingular, and the set of leaders affinely span $\mathbb{R}^d$.
\end{Lemma}
Note that we have used $p_f$ and  $p_l$ to respectively denote the positions of the followers and leaders, so that $p_f^*$ and  $p_l^*$ denote the target positions of the followers and leaders respectively. Let the tracking error be defined by 

\begin{equation}
	\delta _ {p_f}(t) = p_f(t) - p_f^*(t) = p_f(t) + \Omega_{ff}^{-1} \Omega_{fl}p_l^*(t) ,
	\label{Disageement_equation}
\end{equation}

\noindent
so that the control objective is accomplished by ensuring that $p_f(t) \rightarrow  p_f^*(t) $, i.e. $\delta_{p_f} \rightarrow 0 $, as $ t \rightarrow \infty$.

To ensure the leaders are properly selected and the followers targets are defined, we make the following assumption.

\begin{Assumption} \label{Assumption Affine_Localability}
	Assume that $n+1$ leaders are chosen in such that they span the $\mathbb{R}^d$ space affinely. 
\end{Assumption}


\section{Problem Formulation}	
Consider a networked multi-agent system (MAS) with $n$ agents. Let the position of the $i$th agent be denoted by $x_1,...,x_n \in \mathbb{R}^d$, so that the target position of  $i$th agent's in a time-varying formation can be defined by 
\begin{equation}
	x_i^* (t) = \Theta(t) r_i + b(t),
	\label{affine transform_a}
\end{equation}
where $b(t) \in \mathbb{R}^d$ and $\Theta(t) \in \mathbb{R} ^{d \times d}$ are both time-varying. Note that the nominal (reference) configuration has been denoted by $r_i \in \mathbb{R}^d$. Also note that a trivial way of carrying out affine formation control is to change the $\Theta(t)$ and $b(t)$ values for the entire agents (both follower and leader agents) simultaneously. To do this, all the agents would need to have an advance knowledge of the required $\Theta(t)$ and $b(t)$ values for every considered time. This could limit the ability of the agents to respond to a sudden new development, e.g., respond to an obstacle that shows up on their path. Equation (\ref{affine transform_a}) can be rewritten in global form (for all agents) as
\begin{equation}
	x^* (t) = [I_n \otimes \Theta(t)] r + \textbf{1}_n \otimes b(t).
	\label{global affine transform_a}
\end{equation}
where $r=[r_1^T,...,r^T_n] \in \mathbb{R} ^{nd}$ and $x^*(t) \in \mathbb{R} ^{nd}$ have been used to respectively denote the reference configuration and targets to be tracked. The affine image denotes the set of all the affine transformations of the given nominal configuration. Note that all the tracked targets are affine images of the reference configuration $r$.

We define the affine image as a collection of all the affine transformation of the reference configuration $r$. Following \cite{syr22}, the affine image is defined in global form by 
\begin{align}
	\mathcal{A}(r) = \{ &x \in \mathbb{R}^{dn} :~ x = (I_n \otimes \Theta)r + \textbf{1}_n \otimes b, \Theta \in \mathbb{R}^{d \times d},\nonumber \\
	& b \in \mathbb{R}^d   \}. 
	\label{affine image}
\end{align}
Therefore, the overall aim is to find conditions that ensures that $$\lim_{k\to\infty} \parallel x_f(k) - x^*_f(k) \parallel = 0 , ~ \forall x^*_f(k) \in \mathcal{A}(r).$$

in the equivalent discrete-time setting, with only a small subset of the agents (the set of leaders) required to have knowledge of the $\Theta(t)$ and $b(t)$ values in advance.

\subsection{Preamble}
Consider a multi-agent system composed of $n$ agents. Assume that all the agents have continuous-time dynamics, but sense their neighbouring agents at discrete sampling time intervals and all have zero-order-based control inputs. Such that,

\begin{equation*}
	u_{i(t)} = u_{i[k]}, \quad kT\leq t < (k+1)T,
\end{equation*} 

\noindent
where $u_{i(t)},~ k, ~u_{i[k]} ~\mbox{and} ~ T$ respectively denote the control input at $t$ time (continuous-time), discrete-time index, control input at $t=kT$ and sampling period. The following Lemmas are useful in this study.

\begin{Lemma} \cite{me_acc,me_us} \label{lemma linear bilinear transformation}
	The polynomial 
	\begin{equation} \
		s + a = 0,
		\label{linear bilinear transformation}
	\end{equation}
	where $a\in \mathbb{C}$, has its root within a unit circle if  \begin{equation}
		(a+1)t - (a-1)= 0
		\label{proof linear bilinear transformation}
	\end{equation}
	has its roots in the open left half plane.
\end{Lemma}

\begin{Proof}
	The proof directly follows from \cite{me_acc,me_us}, and for convenience, it is reproduced here.
	Using the bilinear transformation $s= \frac{t+1}{t-1}$ \cite{wr_bileaner_transform}, (\ref{linear bilinear transformation}) can be rewritten as
	\begin{equation}
		(a+1)t - (a-1)= 0
		\label{linear bilinear transformation 2}
	\end{equation}
	Note that using bilinear transformation, there exist a one-to-one mapping of the open left half plane of a polynomial to the interior of a unit circle  
\end{Proof}

We assume that the multi-agent system is composed of $n_l$ leaders and $n_f$ followers and begin our study with the case were the agents are modelled using single-integrator dynamics.

\section{Network of Integrator Cases}

Consider the multi-agent system where all agents are modelled with single-integrator agent dynamics described respectively in the continuous- and discrete-time by 
\begin{equation}
	\dot{x}_{i(t)} = u_{i(t)}, \quad i=1,...,n
	\label{continuous_time 1st order}
\end{equation}
and
\begin{equation}
	x_{i(k+1)} = x_{i(k)} + Tu_{i(k)}, \quad i=1,...,n,
	\label{discrete_time 1st order}
\end{equation}
where $x_i$ is used to denote the state of the $i$th agent. 
Note that \eqref{discrete_time 1st order} can be defined in global form for all the agents as
\begin{equation*}
	x_{(k+1)} = x_{(k)} + Tu_{(k)},
	\label{discrete_time 1st order all agents}
\end{equation*}
and for the subset of follower agents as
\begin{equation}
	x_{f(k+1)} = x_{f(k)} + Tu_{f(k)}. \quad \forall i \in f
	\label{discrete_time 1st order follower agents}
\end{equation}
Note that $x ~=~ [x_l^T, ~x_f^T]^T$, and also that the first $d+1$ agents denote the set of leaders. The states of the leader are denoted by $x_l^T$. The other agents are denoted as followers. Their states are denoted by $x_f^T$.

\noindent
Next, we consider the sampled-data formation control cases where the leaders are stationary and dynamic.

\subsection{Stationary Leaders}

Here, the leaders are stationary, i.e. $x_{i(k+1)} = x_{i(k)}, ~ \forall i \in \mathcal{V}_l $. For the continuous-time setting, this is given by $\dot{x}_i = 0, ~ \forall i \in \mathcal{V}_l$. Note that $\mathcal{V}_l$ and $\mathcal{V}_f$ respectively denote the sets of leaders and followers. The sampled-data formation control problem is studied using the protocol 

\begin{equation}
	u_{i(k)} = - \sum w_{ij} (x_{i(k)} - x_{j(k)}), \quad i \in \mathcal{V}_f.
	\label{stationary control protocol}
\end{equation}
Equation \eqref{stationary control protocol} can be rewritten in global form for all the agents as 

\begin{equation}
	u_{(k)} = - (\Omega \otimes I_d)x_{(k)}.
	\label{stationary control protocol global all agent form}
\end{equation}
The stress matrix is partitioned for convenience as 
\begin{equation}
	\Omega =  \begin{bmatrix}
		\Omega_{ll} & \Omega_{lf}\\
		\Omega_{fl} & \Omega_{ff}
	\end{bmatrix}.
	\label{stress matrix partitioning 2}
\end{equation}
Therefore, we can write \eqref{stationary control protocol global all agent form} for only the followers as
\begin{equation}
	u_{f(k)} = - (\Omega_{fl} \otimes I_d)x^*_{l(k)} ~~-~~ (\Omega_{ff} \otimes I_d)x_{f(k)}.
	\label{stationary control protocol global all followers}
\end{equation}

\noindent
Using equation (\ref{stationary control protocol global all followers}), equation (\ref{discrete_time 1st order follower agents}) can be rewritten in matrix form for all the followers as
\begin{equation}
	x_{f(k+1)} = [I_{dn_f} - T (\Omega_{ff} \otimes I_d)] x_{f(k)} - T(\Omega_{fl} \otimes I_d)x^*_{l(k)}
	\label{control law 1 equation}
\end{equation}
\begin{Theorem} \label{theorem_stionar_leader}
	Assume that the agents communication graph is undirected and the leaders are stationary, i.e. $x_{i[k]} = x_{i[k+1]} ~ \forall i \in \mathbb{V}_l$. Also assume that Assumptions $1 -3$ hold (that is, the framework of the agents is universally rigid, so that the the stress matrix has the rank $rank(\Omega) = n-d-1$; and the $d+1$ leaders have been selected such that they affinely span the $\mathbb{R}^d$ space). Let the $i$th  eigenvalue of $- \Omega_{ff}$ be denoted by $\mu_i$. Then, by choosing $T\mu_{\min} > -2$, the control law, equation  (\ref{control law 1 equation}), guarantees that each respective follower agent would stabilize to the desired targets. 	
\end{Theorem}

\begin{Proof}
	Define the global disagreement for all followers by
	\begin{align}
		\delta_{x_f[k+1]} &= x_{f[k+1]} - x_{f[k+1]}^*  \nonumber  \\
		&= x_{f[k+1]} - [-(\Omega_{ff}^{-1}\Omega_{fl} \otimes I_d)]x_{l[k+1]}^* \nonumber \\
		&= x_{f[k+1]} + (\Omega_{ff}^{-1}\Omega_{fl} \otimes I_d)x_{l[k+1]}^*
		\label{disagreement function for delta k+1}
	\end{align}	
	Similarly, 
	\begin{equation}
		\delta_{x_f[k]} = x_{f[k]} + (\Omega_{ff}^{-1}\Omega_{fl} \otimes I_d)x_{l[k]}^*.
		\label{disagreement function for delta k}
	\end{equation}
	By substituting for $x_{f[k+1]}$ in \eqref{disagreement function for delta k+1} using equation \eqref{control law 1 equation} and noting that $x_{l[k+1]}  =  x_{l[k]}  =  x_{l^*[k]}$ (because the leaders are at their target positions already), the expression 
	
	\begin{align}
		\delta_{x_{f[k+1]}} &= x_{f[k]} - T (\Omega_{ff} \otimes I_d) x_{f[k]} - T(\Omega_{fl} \otimes I_d)x^*_{l[k]}  \nonumber \\
		& ~~~+ (\Omega_{ff}^{-1}\Omega_{fl}  \otimes I_d) x_{l[k]}^*   \nonumber \\
		&=-T(\Omega_{ff} \otimes I_d) [x_{f[k]}  ~ + ~ (\Omega_{ff}^{-1}\Omega_{fl}  \otimes I_d) x_{l[k]}^*]  \nonumber \\ & ~~~ +  ~ [x_{f[k]}   + (\Omega_{ff}^{-1}\Omega_{fl}  \otimes I_d) x_{l[k]}^*]    \nonumber \\
		&=-T(\Omega_{ff} \otimes I_d) \delta_{f[k]}  ~ +  ~ \delta_{f[k]} \nonumber \\ 
		&=~  [(-T \Omega_{ff} + I_{n_f}) \otimes I_d] \delta_{f[k]},
		\label{disagreement for control law 1}
	\end{align}
	is obtained.
	Note that equation \eqref{disagreement function for delta k} has been used to obtain equation \eqref{disagreement for control law 1}.

	\noindent
	To guarantee that the followers track their respective targets,  equation (\ref{disagreement for control law 1}) should be stabilized to the origin. Here, this is solved if the term $(-T \Omega_{ff} + I_{n_f})$ have negative eigenvalues. The corresponding characteristic polynomial is defined by $det(sI_{nf} - I_{nf} + T \Omega_{ff})$. Let $\mu_i$ denote the $i$th eigenvalue of $-\Omega_{ff}$. Therefore,
	\begin{equation*}
		det(sI_{nf} - I_{nf} + T \Omega_{ff})) = \Pi^{n}_{i=1} (s - 1 - T\mu_i)
	\end{equation*} 
	This follows that
	\begin{equation}
		s - 1 - T\mu_i = 0.
		\label{stability of the error}
	\end{equation}
	By substituting $a$ for $(- 1 - T\mu_i)$ in equation (\ref{stability of the error}) and considering Lemma (\ref{lemma linear bilinear transformation}), we derive
	\begin{equation}
		-T\mu_i t + T\mu_i + 2 = 0.
		\label{BIL 1}
	\end{equation} 
	Since $-T\mu_i > 0$, therefore $ T\mu_i + 2$ is required to be greater than zero so that the roots of equation (\ref{BIL 1}) would be in the left half plane and equation (\ref{stability of the error}) within the unit circle. This requires that $ T\mu_{\min} > -2 $, where $\mu_{\min}$ has been chosen to denote the lower bound.	
\end{Proof}
\noindent
Next, we present another control law capable of dealing with the case where the positions of the leaders can be dynamic.

\subsection{Dynamic Leaders}
The control law proposed in equation (\ref{control law 1 equation}) cannot guarantee that the tracking errors would decrease to zero for systems having dynamic leaders. To address this scenario, we propose a different control law. Consider the algorithm,

\begin{align}
	u_{i[k]} = &-\frac{1}{\gamma} \sum _{j \in \mathcal{N}_i} w_{ij} (x_{i[k]}  -  x_{j[k]} ~ - ~\frac{1}{T}(x_{j[k+1]} ~-~ x_{j[k]})), \nonumber \\ &i \in \mathcal{V}_f,
	\label{dynamic control protocol}
\end{align}
where $\gamma ~=~  \sum _{j \in \mathcal{N}_i} w_{ij}$.  
Substituting for $u_{i[k]}$ in equation \eqref{discrete_time 1st order follower agents} using equation \eqref{dynamic control protocol} and performing algebraic simplification, the expression (closed loop)  

\begin{align}
	&\sum _{j \in \mathcal{N}_i} w_{ij} (x_{i[k+1]} - x_{j[k+1]})  \nonumber \\
	& \qquad  = (1-T)\sum _{j \in \mathcal{N}_i} w_{ij} (x_{i[k]} - x_{j[k]}), ~ i \in \mathcal{V}_f,
	\label{dynamic control protocol closed loop}
\end{align}
is derived.

The stability of control law (\ref{dynamic control protocol closed loop}) is presented as follows.

\begin{Theorem}
	Let the communication graph of the agents be undirected and the framework be universally rigid, such that the stress matrix has the rank, $rank(\Omega) = n-d-1$. Assume that the requirements of Assumptions $1 - 3$ are satisfied. By choosing $T<2$, control law (\ref{dynamic control protocol closed loop}) guarantees that the followers would track their respective targets and the required affine formation control is accomplished.
\end{Theorem}

\begin{Proof}
	Equation (\ref{dynamic control protocol closed loop}) can be rewritten in global form for all agents as 
	\begin{equation}
		(\Omega \otimes I_d)x_{(k+1)} ~=~ (1-T)(\Omega \otimes I_d)x_{(k)}. 
		\label{global discrete dynamics}
	\end{equation}	
	By noting equation \eqref{stress matrix partitioning 2}, equation \eqref{global discrete dynamics} can then be rewritten for the followers as	
	\begin{align*}
		&(\Omega_{fl} \otimes I_d)x_{l(k+1)} ~+~ (\Omega_{ff} \otimes I_d)x_{f(k+1)} \\
		& \quad = (1-T)[(\Omega_{fl} \otimes I_d)x_{l(k+1)} ~+~ (\Omega_{ff} \otimes I_d)x_{f(k+1)}].
	\end{align*}
	Multiplying through by $(\Omega_{ff}^{-1} \otimes I_d)$ (from the left hand side), the expression
	\begin{align}
		&(\Omega_{ff}^{-1} \Omega_{fl} \otimes I_d)x_{l(k+1)} ~+~ x_{f(k+1)} \nonumber \\
		& \quad \qquad = (1-T)[(\Omega_{ff}^{-1} \Omega_{fl} \otimes I_d)x_{l(k+1)} ~+~ x_{f(k+1)}]
		\label{stationary control protocol global all followers 2}
	\end{align}
	is derived.
	Define the error (or disagreement) as $\delta_{x_{f[k]}}= 	x_{f[k]} - x_{f[k]}^* =  x_{f[k]}  + \Omega_{ff}^{-1} \Omega_{fl} x^*_{l[k]} $. Therefore, control law (\ref{stationary control protocol global all followers 2}) can be written in terms of the errors (disagreements) as $\delta_{x_f[k+1]} = (1-T)I_{dn_f} \delta{x_{f[k]}}$. The characteristic equation satisfies $s + T -1 = 0$. Using the bilinear transformation $s=\frac{t+1}{t-1}$, the expression
	\begin{equation*}
		Tt - T + 2 = 0
	\end{equation*}
	is derived.
	Considering that $Tt> 0$, it is therefore required that $-T+2>0$ to satisfy the overall system stability requirement. This is requirement is satisfied by selecting the sampling period $T$, such that $T<2$.
	
	
\end{Proof}


\subsection{Simulation Study}

\begin{figure}[!ht]
	\begin{center}
		\includegraphics[width=0.5\textwidth]{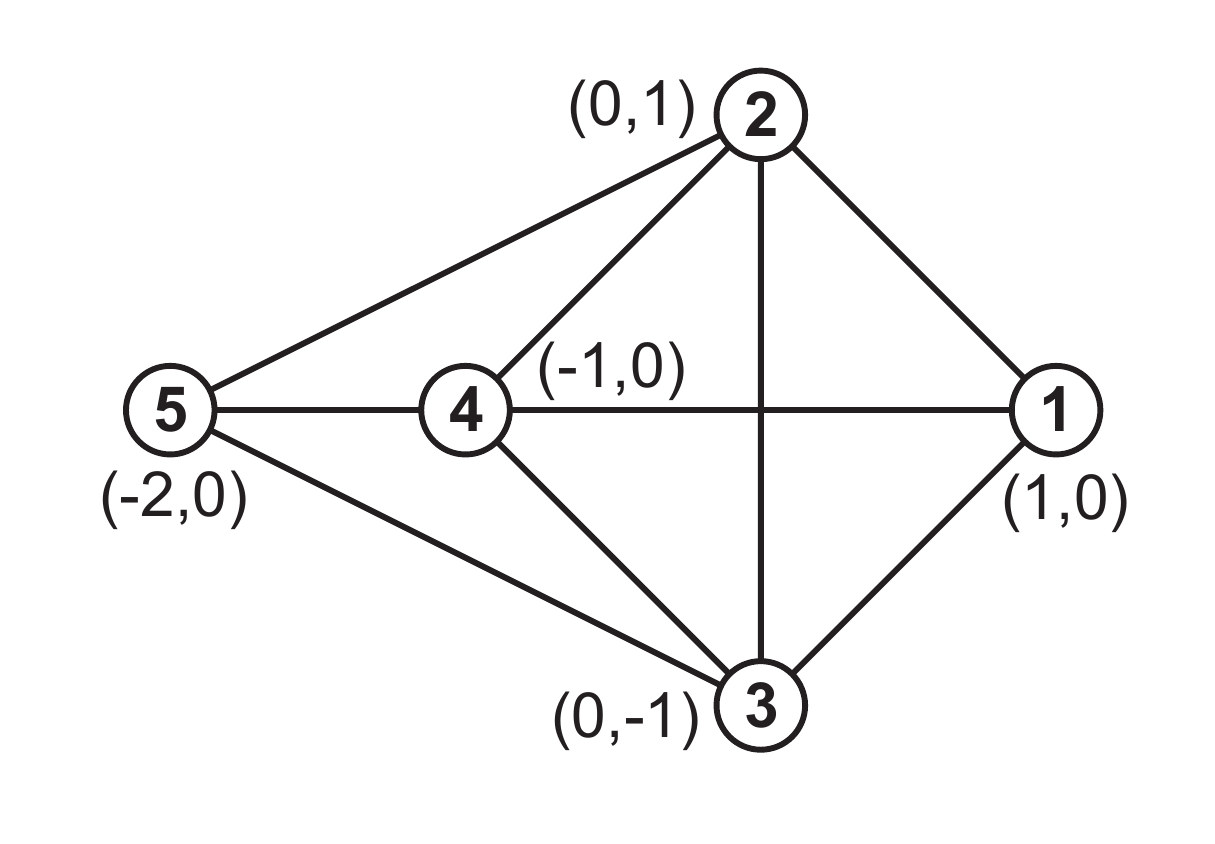}
		\caption{Framework depicting 5 agents communication together with their $2$-dimensional reference positions \cite{me_us,me_icca}. The agents communications are depicted with straight lines.}
		\label{Sample Formation 5 Agents}
	\end{center}
\end{figure}

\begin{figure*}[!ht]
	\begin{center}
		\includegraphics[width=\textwidth]{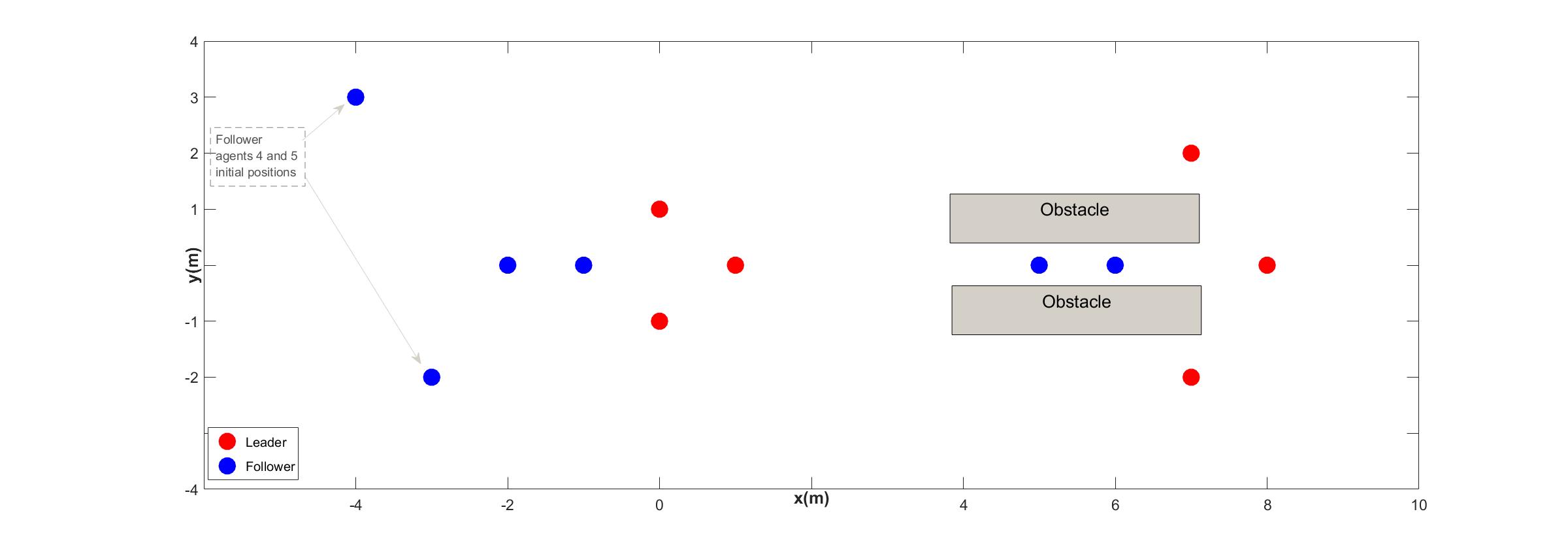}
		\caption{Illustration of required formation control accomplishment.  Here, $\mu_{\min}$ is computed as $-1.49$. Control law \ref{control law 1 equation} is used for the simulation and $T=1$ is chosen. Leader agents $1-3$ were initialized to $P_1(1, 0), ~ P_2(0,1)~ \mbox{and} ~ P_3(0,-1)$ respectively. Similarly, the follower agents $4$ and $5$ were initialized to $P_4(-4,3) ~ \mbox{and} ~ P_5(-3,-2)$ respectively. The simulation was first allowed to run for 3 mins and then the data for the new positions of the agents were obtained. As expected, the followers agents correctly tracked their targets, i.e., agent 4 moved to positions (-1,0) and agent 5 moved to position (-2,0) as the position of the leaders remained unchanged. Clearly, this is the required reference configuration as shown in Figure \eqref{Sample Formation 5 Agents}. After this is accomplished, the position of the leaders were then changed and the followers correctly tracked their their new corresponding positions after the movement of the leaders.}
		\label{formation}
	\end{center}
\end{figure*}

In this study, the formation control of multi-robotic (multi-agent) systems with periodic inter-agent communication is considered.
Consider a multi-agent system consisting of 5 agents each modelled with single integrator agent dynamics. Here, we use control law \eqref{control law 1 equation} and Theorem \eqref{theorem_stionar_leader}. Agents $1$ to $3$ are chosen as leaders and the remaining agents are followers. The Framework, Fig. \eqref{Sample Formation 5 Agents}, is used to depict the agents communication graph and the corresponding stress matrix is given by \cite{me_us,me_icca}
%
%
\begin{equation*}
	\Omega =
	\begin{bmatrix}
		&\Omega_{ll}   &\Omega_{lf}  \\
		&\Omega_{fl}   &\Omega_{ff}  \\
	\end{bmatrix}\\
\end{equation*}
where

\begin{align*}
	\Omega_{ll} &=
	\begin{bmatrix}
		&    0.292  &   -0.292 &  -0.292  \\
		&   -0.292  &    0.354 &   0.354  \\
		&   -0.292  &    0.354 &   0.354
	\end{bmatrix},
	\\
	\Omega_{lf} &=
	\begin{bmatrix}
		&    0.292  &      0  \\
		&   -0.542  &  0.125 \\
		&   -0.542  &  0.125
	\end{bmatrix},
	\\
	\Omega_{ff} &=
	\begin{bmatrix}
		&1.292         &-0.500\\
		&-0.500        &0.250 
	\end{bmatrix},  \quad \text{and}
	\\
	\Omega_{fl} &=
	\begin{bmatrix}
		&0.292    &-0.542   &-0.542\\
		&0        &0.125    &0.125 
	\end{bmatrix}.
\end{align*}

\noindent
The smallest eigenvalue of $-\Omega_{ff}$ is $-1.49$ (i.e., $\mu_{\min}=-1.49$). The sampling period of $1$ (i.e., $T=1$sec) is used. Clearly, this satisfies the $T \mu_{\min} > -2$ requirement of Theorem \eqref{theorem_stionar_leader}.

The study was carried out in a simulation platform using MATLAB software. The simulation result is presented in Fig. \ref{formation}. It shows the agents first produced the required reference configuration with the two follower agents $4$ and $5$ initial positions set to the $P_4(-4, 3)$ and $P_5(-3,-2)$ meter marks respectively. The agents also executed a scaling manoeuvre to avoid an obstacle.
Note that we have used $P_4(-4,3)$ to denote that agent $4$ (i.e., $P_4$) in the $x-y$ plane has its $x$-component as $-4$ and $y$-component as $3$.


\section{General Linear Systems}	


%

Here, we consider the the case where each follower agent (or node) has dynamics described by

\begin{equation}
	x_{i(k+1)} = Ax_{i(k)} ~ +~ Bu_{i(k)},
	\label{local_agent_dynamics}
\end{equation} 
where $A$ and $B$ are constant matrices of appropriate dimensions. Note that the global form of \eqref{local_agent_dynamics} is given by
\begin{equation}
	x_{(k+1)} = (I_n \otimes A)x_{(k)} ~ +~ (I_n \otimes B)u_{(k)},
	\label{global_agent_dynamics}
\end{equation}
Our overall goal remains to find conditions that guarantees that $$\lim_{k\to\infty} \parallel x_f(k) - x^*_f(k) \parallel = 0 , ~ \forall x^*_f(k) \in \mathcal{A}(r).$$
Note that we have used $n, ~n_l ~ \mbox{and} ~n_f$ to denote the number of agents, leader-agents, and follower-agents respectively. Similarly, we have used $x_n, ~x_l ~ \mbox{and} ~x_f$ to denote the state of the agents, leaders, and followers, respectively.

\noindent
Consider the control protocol
\begin{equation}
	u_i = K\sum_{j=1}^{n} [ \epsilon w_{ij} (x_{i(k)} ~-~ x_{j(k)}) ~+~ x_{i(k)} ]
	\label{linear_control_protocol}
\end{equation}
whose global form is given by
\begin{equation}
	u = [(I_{n} - \epsilon \Omega) ~\otimes~ K]x_{(k)}
	\label{global_control_protocol}
\end{equation}
where $K$ is a local feedback gain and $\epsilon$ is a design parameter to help guarantee the system stability. Using \eqref{global_control_protocol} and \eqref{global_agent_dynamics}, the global dynamics can be written as 

\begin{equation}
	x_{(k+1)} = [(I_n \otimes A) ~ +~ (I_{n} - \epsilon \Omega)  \otimes BK)]x_{(k)},
	\label{overall_global_agent_dynamics}
\end{equation}

\noindent
The stability of control law (\ref{overall_global_agent_dynamics}) is presented as follows.

\begin{Theorem}
	Assume that $B$ has full column rank, and the pair $(A,B)$ is stabilizable. Let $Q=Q^T>0$ and $P>0$ be the unique solution to the modified algebraic Riccati equation
	\begin{equation*}
	A^TPA ~-~ P ~-~	A^TPB(B^TPB)^{-1}B^TPA ~+~ Q ~=~ 0.
	\end{equation*}
	Let the communication graph of the agents be undirected and the framework be universally rigid, such that the stress matrix has the rank, $rank(\Omega) = n-d-1$. Assume that the requirements of Assumptions $1 - 3$ are satisfied. By choosing $K=-(B^TPB)^{-1}B^TPA$, control law (\ref{overall_global_agent_dynamics}) solves the affine formation control problem for the follower agents.

\end{Theorem}

\begin{Proof}
	The proof of the theorem is omitted because of page restrictions and plans for a future publication of the expanded version of this work.
\end{Proof}

\section{Conclusion}
In this study, the distributed leader-follower affine formation control of multi-agent systems with periodic communication is considered. The stress-matrix based approach is used for the formation control. Control laws are proposed in the study of different scenarios. Sufficient conditions to guarantee the systems stability are provided. With the proposed control laws tracking of both stationary and dynamic targets can be achieved for formations control manoeuvres such as scaling, translations, rotations, etc. Simulation studies are used to demonstrate the efficacy of the proposed control laws. Ongoing work is to generalize the solution to discrete-time systems with uncertainties.

\reftitle{References}

\bibliography{mybibfile_3}

\end{document}